\documentclass[conference]{IEEEtran}
\IEEEoverridecommandlockouts

\usepackage{cite}
\usepackage{amsmath,amssymb,amsfonts}
\usepackage{algorithmic}
\usepackage{graphicx}
\usepackage{textcomp}
\usepackage{xcolor}

\usepackage[citecolor=blue,colorlinks=true,urlcolor=blue,linkcolor=blue,bookmarks=false]{hyperref}
\usepackage[colorinlistoftodos]{todonotes}
\usepackage{url}
\usepackage{algorithm}
\usepackage{algorithmic}
\usepackage{lipsum}
\usepackage[keeplastbox]{flushend}

\usepackage{booktabs}

\usepackage{subfig}
\title{\LARGE \bf
Deep Neural Network Ensembles for Time Series Classification
}

\author{\IEEEauthorblockN{Hassan Ismail Fawaz,
Germain Forestier,
Jonathan Weber,
Lhassane Idoumghar and 
Pierre-Alain Muller\smallskip }
\IEEEauthorblockA{IRIMAS,
Universit\'e Haute-Alsace, Mulhouse, France \smallskip}
\IEEEauthorblockN{\textit{Email}: \{first-name.last-name@uha.fr\}}
}

\begin{document}
\bstctlcite{IEEEexample:BSTcontrol}
\maketitle
\thispagestyle{empty}
\pagestyle{empty}

\begin{abstract}
Deep neural networks have revolutionized many fields such as computer vision and natural language processing. 
Inspired by this recent success, deep learning started to show promising results for Time Series Classification (TSC). 
However, neural networks are still behind the state-of-the-art TSC algorithms, that are currently composed of ensembles of 37 non deep learning based classifiers. 
We attribute this gap in performance due to the lack of neural network ensembles for TSC. 
Therefore in this paper, we show how an ensemble of 60 deep learning models can significantly improve upon the current state-of-the-art performance of neural networks for TSC, when evaluated over the UCR/UEA archive: the largest publicly available benchmark for time series analysis.
Finally, we show how our proposed Neural Network Ensemble (NNE) is the first time series classifier to outperform COTE while reaching similar performance to the current state-of-the-art ensemble HIVE-COTE. 
\end{abstract}

\section{Introduction}

Time series data are omnipresent in many practical data science applications ranging from health care~\cite{IsmailFawaz2018evaluating} and stock market predictions~\cite{anghinoni2018time} to social media analysis~\cite{xu2018mnrd} and human activity recognition~\cite{xi2018deep}.  
Since 2006, time series analysis has been considered one of the most challenging problems in data mining~\cite{yang200610}, and in a more recent poll it has been shown that 48\% of data expert had analyzed time series data during their career, ahead of text and images~\cite{piatetsky2014}. 

Time Series Classification (TSC) tasks differ from traditional classification tasks by the natural temporal ordering of their attributes~\cite{bagnall2017the}. 
To tackle this problem, a huge amount of research was dedicated into coupling and enhancing time series similarity measures with a Nearest Neighbor (NN) classifier~\cite{dau2017judicious,gharghabi2018ultra}. 
In~\cite{lines2015time}, ten elastic distances were compared to the traditional Dynamic Time Warping (DTW) algorithm to find out that no single measure could outperform the classic NN coupled with DTW (NN-DTW) for TSC.
These findings motivated the authors to construct a single Elastic Ensemble (EE) classifier that includes all eleven different similarity measures, and achieve a significant improvement compared to the individual classifiers~\cite{lines2015time}. 
Hence, recent contributions were focused on ensembling different discriminant classifiers such as decision trees (random forest)~\cite{baydogan2013a} and Support Vector Machines (SVMs)~\cite{bostrom2015binary} on different data representation techniques such shapelet transform~\cite{bostrom2015binary} or DTW features~\cite{kate2016using}. 
These ideas gave rise to the Collective Of Transformation-based Ensembles (COTE)~\cite{bagnall2016time} and its extended version HIVE-COTE~\cite{lines2018time} where 37 different classifiers were ensembled over multiple time series data transformation techniques in order to reach current state-of-the-art performance for TSC~\cite{bagnall2017the}. 

With the advent of deep neural networks into industrial and commercial applications such as self-driving cars~\cite{qiu2018multi} and speech recognition systems~\cite{liul2018stochastic}, time series data mining practitioners started investigating the application of deep learning to TSC problems~\cite{wang2017time}.
In our recent empirical study~\cite{IsmailFawaz2018deep}, we showed how deep Convolutional Neural Networks (CNNs) are able to achieve results that are not significantly different than current state-of-the-art algorithms for TSC problems when evaluatedover the 85 time series datasets from the UCR/UEA archive~\cite{ucrarchive,bagnall2017the}.   
Outside the UCR/UEA benchmark, deep neural networks have seen some very successful applications such as evaluating surgical skills from multivariate time series~\cite{IsmailFawaz2018evaluating} and recognizing human activities from wearable sensors data~\cite{xi2018deep}. 
These results suggest that building upon deep learning based solutions for TSC could further improve the current state-of-the-art performance of deep neural networks. 

One way of improving neural network based classifiers is to build an ensemble of deep learning models. 
This idea seems very interesting for TSC tasks since the state-of-the-art is moving towards ensembled solutions~\cite{lines2018time,lines2015time,bagnall2017the,baydogan2013a}. 
In addition, deep neural network ensembles seem to achieve very promising results in many supervised machine learning domains such as skin lesions detection~\cite{goyal2018deep}, facial expression recognition~\cite{wen2017ensemble} and automatic bucket filling~\cite{dadhich2018predicting}. 

\begin{figure}
    \centering
    \includegraphics[width=0.95\linewidth]{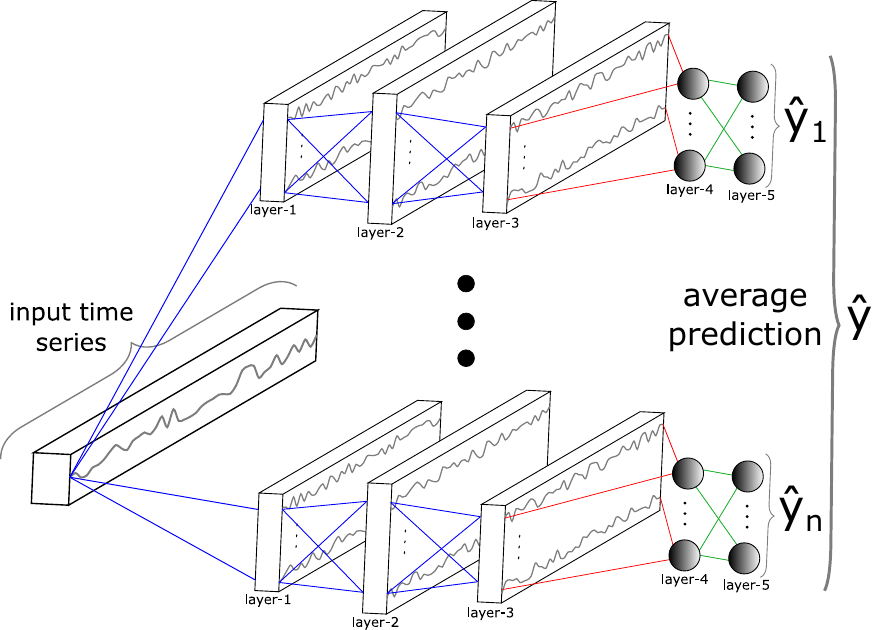}
    \caption{Ensemble of deep convolutional neural networks for time series classification.}
    \label{fig:ensemble}
\end{figure}

Therefore, we propose to ensemble the current state-of-the-art deep learning models for TSC developed in~\cite{IsmailFawaz2018deep}, by constructing one model composed of 60 different deep neural networks: 6 different architectures~\cite{wang2017time,zheng2014time,zhao2017convolutional,serra2018towards} each one with 10 different initial weight values.
By evaluating on the 85 datasets from the UCR/UEA archive, we demonstrate a significant improvement over the individual classifiers while also reaching very similar performance to HIVE-COTE: the current state-of-the-art ensemble of 37 non deep learning based time series classifiers.  
Finally, inspired by the recent success of transfer learning for TSC~\cite{IsmailFawaz2018transfer}, we replace ensembling randomly initialized networks with an ensemble constructed out of fine-tuned models from 84 different source datasets, which showed a significant improvement for TSC problems.

The paper is divided as follows, we first start by explaining the background material, before presenting our different techniques of ensembling deep neural networks. 
We then describe our results and discussions before drafting a final conclusion with our future directions.   

\section{Background}
In this section we describe the current state of research in neural networks for TSC and then present some work related to ensembling neural network classifiers.    

\subsection{Neural networks for time series classification}
Since AlexNet~\cite{krizhevsky2012imagenet} won the ImageNet~\cite{russakovsky2015imagenet} competition in 2012 with a significant improvement in accuracy compared to previous  state-of-the-art approaches, the computer vision field was revolutionized with many deep neural networks papers being published every year to solve image recognition and object localization problems~\cite{lecun2015deep}.     
In addition, sequential data mining tasks such as natural language processing and speech recognition are being tackled with deep convolutional, recurrent and generative adversarial neural networks~\cite{li2018text,wang2018hierarchical}. 

Inspired by this recent success of deep learning models, researchers started adopting these complex machine learning techniques to solve the underlying task of Time Series Classification~\cite{IsmailFawaz2018deep,IsmailFawaz2019adversarial}. 
Specifically Wang \textit{et al.}~\cite{wang2017time} showed very promising results, where a Fully Convolutional Neural network (FCN) and a Residual Network (ResNet) were designed to reach COTE's performance when evaluated on 44 datasets from the UCR/UEA archive~\cite{ucrarchive,bagnall2017the}. 
Moreover, in our recent empirical evaluation of deep learning models for TSC~\cite{IsmailFawaz2018deep}, we managed to reinforce these findings by testing FCN and ResNet on 85 datasets from the UCR/UEA archive.  
In fact, similar to two dimensional data (images), one dimensional convolutions when slid over an input time series, enable a non-linear transformation of the data. 
By applying backpropagation over a cascade of several convolutional layers with many filters, the network is able to learn this time invariant hierarchical representation of the input time series which is potentially useful for classification. 
For more detail about how these convolutions are being applied to one dimensional time series data, we refer the interested reader to our recent survey of deep learning for time series classification~\cite{IsmailFawaz2018deep}. 

Different variants of CNNs were proposed for TSC and validated on the UCR/UEA archive. 
Multi-scale CNN (MCNN)~\cite{cui2016multi} was among the first deep learning architectures to be evaluated for domain agnostic TSC.  
In~\cite{leguennec2016data} Time LeNet (t-LeNet) was proposed as an adaptation of the famous LeNet architecture which was originally proposed for document recognition~\cite{lecun1998gradient}.
Multi-Channels Deep Convolutional Neural Networks (MCDCNN)~\cite{zheng2014time} and Time-CNN~\cite{zhao2017convolutional} were originally proposed for multivariate TSC, however in~\cite{IsmailFawaz2018deep} we have shown how they can be easily extended for univariate TSC.
One last CNN model called Encoder was proposed in~\cite{serra2018towards} where FCN was extended to include the attention mechanism. 
Adding to the aforementioned neural network architectures, the classical Multi-Layer Perceptron (MLP) was considered as a baseline architecture in~\cite{wang2017time}.
Finally, we should mention in addition to this pool of deep CNNs for TSC, a non-convolutional recurrent model called Time Warping Invariant Echo State Networks (TWIESN)~\cite{tanisaro2016time}, which showed promising results on different datasets in the archive~\cite{IsmailFawaz2018deep}. 

In~\cite{IsmailFawaz2018deep}, we showed how ResNet, FCN and Encoder won on 43, 18 and 10 datasets respectively suggesting that indeed no single network would outperform all the others on the whole benchmark.
This would motivate researchers to ensemble the decision of these deep learning classifiers, which is the main contribution of this paper: showing how an ensemble of different deep neural networks can outperform all single individual classifiers and reach new state-of-the-art performance for TSC.   

\subsection{Neural networks ensemble}
Constructing an ensemble of many deep learning classifiers has been shown to achieve high performance in many different fields. 
In~\cite{goyal2018deep}, an ensemble of two neural networks was adopted: (1) Inception-v4 and (2) Inception-ResNet-v2. 
Both of these classifiers are learned with a joint meta-learning approach in an end-to-end manner.
A forest CNN was proposed in~\cite{lee2017ensemble} for image classification, where similarly to random forest, the ensemble is constructed by replacing the individual nodes with a CNN and finally the classifier's decision is taken by performing a majority voting scheme over the different decisions of the individual trees in the forest. 
Another ensemble of CNNs for facial expression recognition was proposed in~\cite{wen2017ensemble} where each individual classifier was trained independently to output a probability for each class and then the network's final decision was taken using a probability-based fusion method. 
In~\cite{dadhich2018predicting}, an ensemble of neural networks was found to outperform other hybrid machine learning ensembles when solving an automatic bucket filling problem. 
Finally in~\cite{ienco2018semi}, deep auto-encoders were ensembled in order to learn an unsupervised latent representation of the input data over multiple resolutions, thus improving the quality of the produced clusters.

Although in almost all use cases ensembling deep neural networks almost always yields to better decisions, we did not find any approach using a neural network ensemble for domain agnostic TSC.
Perhaps the work in~\cite{jin2016ensemble} is the closest to ours where a neural network based ensemble was used to perform biomedical TSC, where individual architectures were constructed with some domain knowledge specific to the classification problem at hand such as choosing the filter length with local and distorted views. 
In addition, our recent work on ensembling two deep learning models (with or without data augmentation) showed how the ensemble classifier was able to outperform significantly the individual model~\cite{IsmailFawaz2018data}. 
Therefore, we decided to further explore ensembling deep neural networks for TSC, by combining multiple deep learning models in different settings. 

\section{Methods}\label{sec:methods}
In this section, we start by presenting the six different architectures composing our ensembles of neural networks. 
For completeness, we describe the random initialization technique adopted for all models. 
Finally, we present a transfer learning based alternative to randomly initializing the weights of the networks. 

\subsection{Architectures}

The average rank of the six chosen deep learning classifiers, over the 85 datasets from the UCR/UEA archive~\cite{ucrarchive,bagnall2017the} is listed in Table~\ref{tab:avg-rank}. 
All of these architectures were implemented in a common framework during our empirical study~\cite{IsmailFawaz2018deep}, containing originally  9 different deep learning approaches for TSC. 
However only 6 out of these 9 approaches were probabilistic classifiers whereas the three other classifiers performed a hard prediction: meaning an input time series is assigned a specific class rather than a probability distribution over all the classes in a dataset.    
Therefore, we chose to only ensemble the 6 probabilistic models, thus allowing us to combine the networks by averaging the a posteriori probability for each class over the individual classifiers' output.  
Finally, we present a brief description of these 6 different architectures and refer the interested reader to a more thorough explanation in the corresponding papers. 
All hyperparameters can be found in~\cite{IsmailFawaz2018deep}.

\begin{table}
    \centering
    \begin{tabular}{|c|c|c|}
    \toprule
         \textbf{Approach} & \textbf{Rank} & \textbf{Wins} \\
         \midrule
         ResNet~\cite{wang2017time} & 1.88 & 41 \\
         FCN~\cite{wang2017time} & 2.49 & 18 \\
         Encoder~\cite{serra2018towards} & 3.34 & 10 \\
         MLP~\cite{wang2017time} & 4.08 & 4 \\
         Time-CNN~\cite{zhao2017convolutional} & 4.38 & 4 \\
         MCDCNN~\cite{zheng2014time} & 4.83 & 3 \\
         \bottomrule
    \end{tabular}
    \caption{Average rank of the six classifiers constituting the Neural Network Ensemble for time series classification over the 85 datasets from the UCR/UEA archive.}
    \label{tab:avg-rank}
\end{table}

\subsubsection{Multi-Layer Perceptron}
(MLP) is the simplest form of deep neural networks and was proposed in~\cite{wang2017time} as a baseline architecture for TSC. 
The architecture contains three hidden layers, with each one fully connected to the output of its previous layer.  
The main characteristic of this architecture is the use of a Dropout layer~\cite{srivastava2014a} to reduce overfitting. 
One disadvantage is that since the input time series is fully connected to the first hidden layer, the temporal information in a time series is lost~\cite{IsmailFawaz2018deep}.  

\subsubsection{Fully Convolutional Neural Network}
(FCN), originally proposed in~\cite{wang2017time}, is considered a competitive architecture yielding the second best results when evaluated on the UCR/UEA archive (see Table~\ref{tab:avg-rank}).
This network is comprised of three convolutional layers, each one performing a non-linear transformation of the input time series. 
A global average pooling operation is used before the final softmax classifier, thus reducing drastically the number of parameters in a network and allowing an architecture that is invariant to the length of the input time series. 
The latter characteristic motivated us to perform a transfer learning technique in~\cite{IsmailFawaz2018transfer}, and ensembling the resulting neural networks which is later discussed in Section~\ref{sec:transfer}.


\subsubsection{Residual Network}
(ResNet) was originally proposed in~\cite{wang2017time} and showed similar performance to FCN when evaluated on 44 datasets from the archive. 
However, when evaluated over the 85 datasets, ResNet significantly outperformed FCN (see Table~\ref{tab:avg-rank}). 
The main characteristic of ResNet is the addition of residual connections which enables a direct flow of the gradient~\cite{wang2017time}.

\subsubsection{Encoder}
(Encoder) was originally proposed in~\cite{serra2018towards} as a hybrid CNN that modifies the FCN architecture~\cite{wang2017time} by mainly adding a Dropout layer~\cite{srivastava2014a} and an attention mechanism. 
The latter operation enables Encoder to learn to localize which regions of the input time series are useful for a certain class identification. 

\subsubsection{Multi-Channels Deep Convolutional Neural Networks}
(MCDCNN) was originally proposed in~\cite{zheng2014time} for multivariate TSC and adapted to univariate data in~\cite{IsmailFawaz2018deep}.
It consists of a traditional CNN, where each convolutional layer is followed by a max pooling operation, then a traditional fully connected layer is used before the final softmax classifier. 

\subsubsection{Time Convolutional Neural Network}
(Time-CNN) was originally proposed for univariate as well as multivariate TSC~\cite{zhao2017convolutional}. 
Similarly to MCDCNN, this network is a traditional CNN with one major exception: the use of the mean squared error instead of the traditional categorical cross-entropy loss function, which has been used by all the deep learning approaches we have mentioned so far.
Therefore for Time-CNN, the sum of the output class probabilities is not guaranteed to be equal to one.  

\subsection{Ensembling models with random initial weights}
We have described in the previous subsection, the architecture of six different classifiers. 
The weights for each network are initialized randomly using Glorot's uniform initialization method~\cite{glorot2010understanding}. 
This technique ensures a uniform distribution of the initial weight values.
However due to non-convexity, networks with the same architecture but different initial weights could yield different validation accuracy. 
In~\cite{choromanska2015loss}, the authors showed that deeper networks are much more stable with respect to the randomness.
This would suggest that ensembling relatively non deep architectures would yield to a much better improvement in accuracy than ensembling deeper architectures. 
Fortunately, for low dimensional time series data, current state-of-the-art architectures are much less deeper than their counterpart networks for high dimensional images. 
Therefore, we believe that we can leverage this instability of neural networks for time series data by ensembling the decision taken by the same network but with different random initializations, using the following equation: 
\begin{equation}
    \hat{y}_{i,c}=\frac{1}{n}\sum_{j=1}^{n}\sigma_c(x_i,\theta_j) ~~|~~\forall c\in [1,C]
\end{equation}
with $\hat{y}_{i,c}$ denoting the ensemble's output probability of having the input time series $x_i$ belonging to class $c$, which is equal to the logistic output $\sigma_c$ averaged over the $n$ randomly initialized models.   
We should note that training an ensemble of the same architecture with different initial weight values has been shown to improve neural network's performance on many computer vision problems~\cite{wen2017ensemble}, however, we did not encounter any previous work that combines such classifiers for TSC.  

\subsection{Transfer learning}\label{sec:transfer}
An alternative to training a deep classifier from scratch is to fine-tune a model that has been already pre-trained on a un/related task~\cite{IsmailFawaz2018transfer}. 
This process is called transfer learning, where the network is first trained on a source dataset, then the final layer is removed and replaced with a new randomly initialized softmax layer whose number of neurons is equal to the number of classes in the target dataset. 
The pre-trained model is then fine-tuned or re-trained on the target dataset's training set. 
With 85 datasets in the archive, each target dataset will have 84 potential source datasets, which motivated us to ensemble the decision of these 84 FCN models.

\section{Results}
In this section we present the results of different ensembling schemes when evaluated on the 85 datasets from the UCR/UEA archive~\cite{ucrarchive,bagnall2017the}, which is currently the largest publicly available benchmark for time series analysis.
In order to compare multiple classifiers over several datasets, following the recommendations in~\cite{demsar2006statistical}, we perform the Friedman test to first reject the null hypothesis. 
For the post-hoc analysis, following the recent recommendations in~\cite{benavoli2016should}, we abandoned the average rank comparison in favor of a pairwise statistical comparison: the Wilcoxon signed-rank test with Holm's alpha correction ($\alpha=5\%$). 
Finally, we used a critical difference diagram~\cite{demsar2006statistical} to visualize the results of these statistical tests projected onto the average rank axis, with a thick horizontal line showing a clique of classifiers that are not significantly different (see Figure~\ref{fig:cd-diagram-resnets} for an example of such diagram).
All experiments were conducted on a hybrid cluster of more than 60 NVIDIA GPUs comprised of GTX 1080 Ti, Tesla K20, K40 and K80.
Note that the code, the raw results and all the pre-trained models are publicly available on the paper's companion repository\footnote{\url{https://github.com/hfawaz/ijcnn19ensemble}}.

\subsection{Ensembling randomly initialized models}

\begin{figure}
    \centering
    \includegraphics[width=\linewidth]{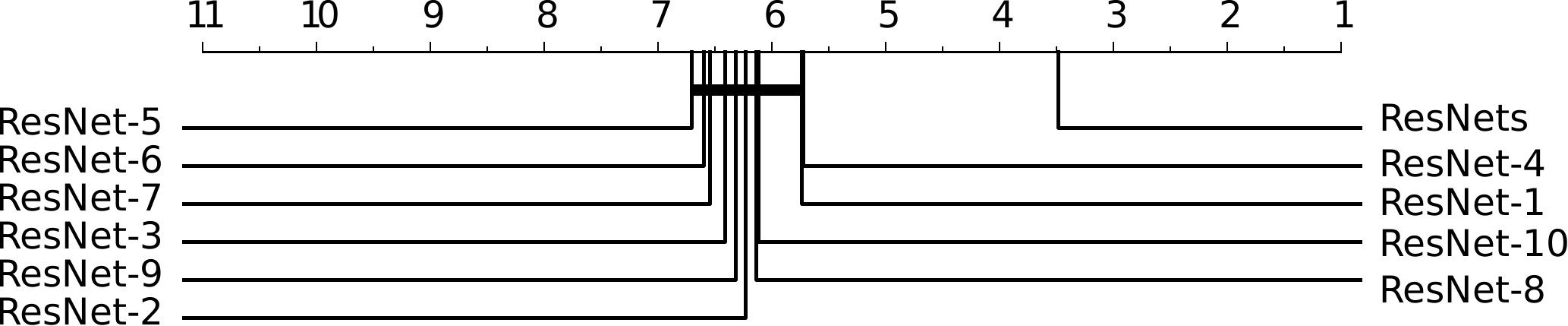}
    \caption{Critical difference diagram showing the pairwise statistical comparison of ten ResNets with random initializations as well as one ResNet ensemble composed of these ten individual neural networks.}
    \label{fig:cd-diagram-resnets}
\end{figure}

By ensembling randomly initialized networks, we are able to achieve a significant improvement in accuracy. 
Figure~\ref{fig:cd-diagram-resnets} shows a critical difference diagram where ten different random initializations of ResNet did not yield to significantly different results.
However, by ensembling these different networks, we were able to demonstrate a significant improvement in the average rank over the 85 datasets.
We should note that the latter phenomenon was also observed for the five other neural networks described in Section~\ref{sec:methods}. 
Finally, we should emphasize that an ensembling technique will improve the stability of ResNet in terms of accuracy, in other words reducing the bias due to the initial weight values as well as the randomness induced by gradient descent based optimization. 

\subsection{Ensembling all neural networks}

\begin{figure}
    \centering
    \includegraphics[width=\linewidth]{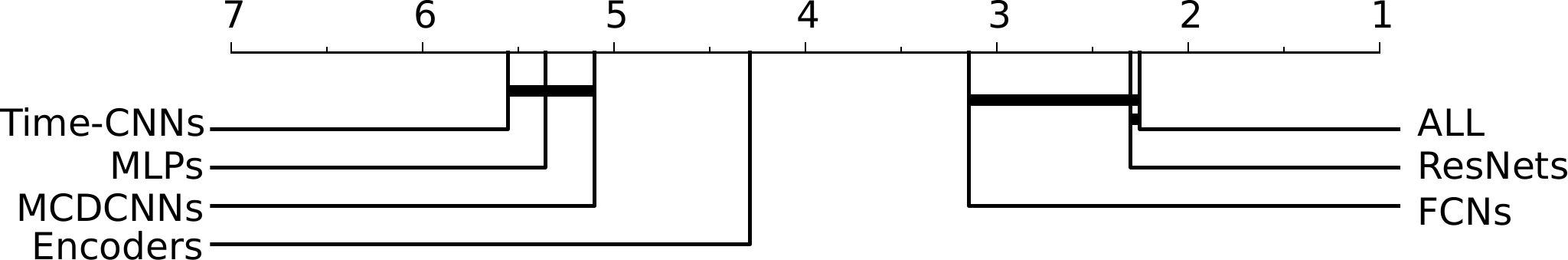}
    \caption{Critical difference diagram showing the pairwise statistical comparison of six architectures ensembled with ten different random initializations each, as well as one ensemble containing the six models.}
    \label{fig:cd-diagram-all}
\end{figure}

After demonstrating that using an ensemble of neural networks is always better than a single classifier, we sought to answer the following question: \textit{Could an ensemble of hybrid randomly initialized networks achieve even better performance?}
Figure~\ref{fig:cd-diagram-all} shows a critical difference diagram containing six ensembles of homogenized networks as well as the hybrid ensemble of \emph{all} available networks. 
The latter classifier contains sixty different networks: each architecture (six in total) is initialized with ten different random weight values.  
The results show that ensembling all networks was able to outperform all classifiers. 
However the statistical test failed to find any significant difference between the full ensemble and individual ResNet/FCN ensembles. 
This would suggest that the ensemble is highly affected by the poor performance of Time-CNN, MLP and MCDCNN. 
The latter classifiers showed the worst average rank without any significant difference, thus suggesting that removing them would yield even better performance.

\subsection{Neural Network Ensemble}

\begin{figure}
    \centering
    \includegraphics[width=0.8\linewidth]{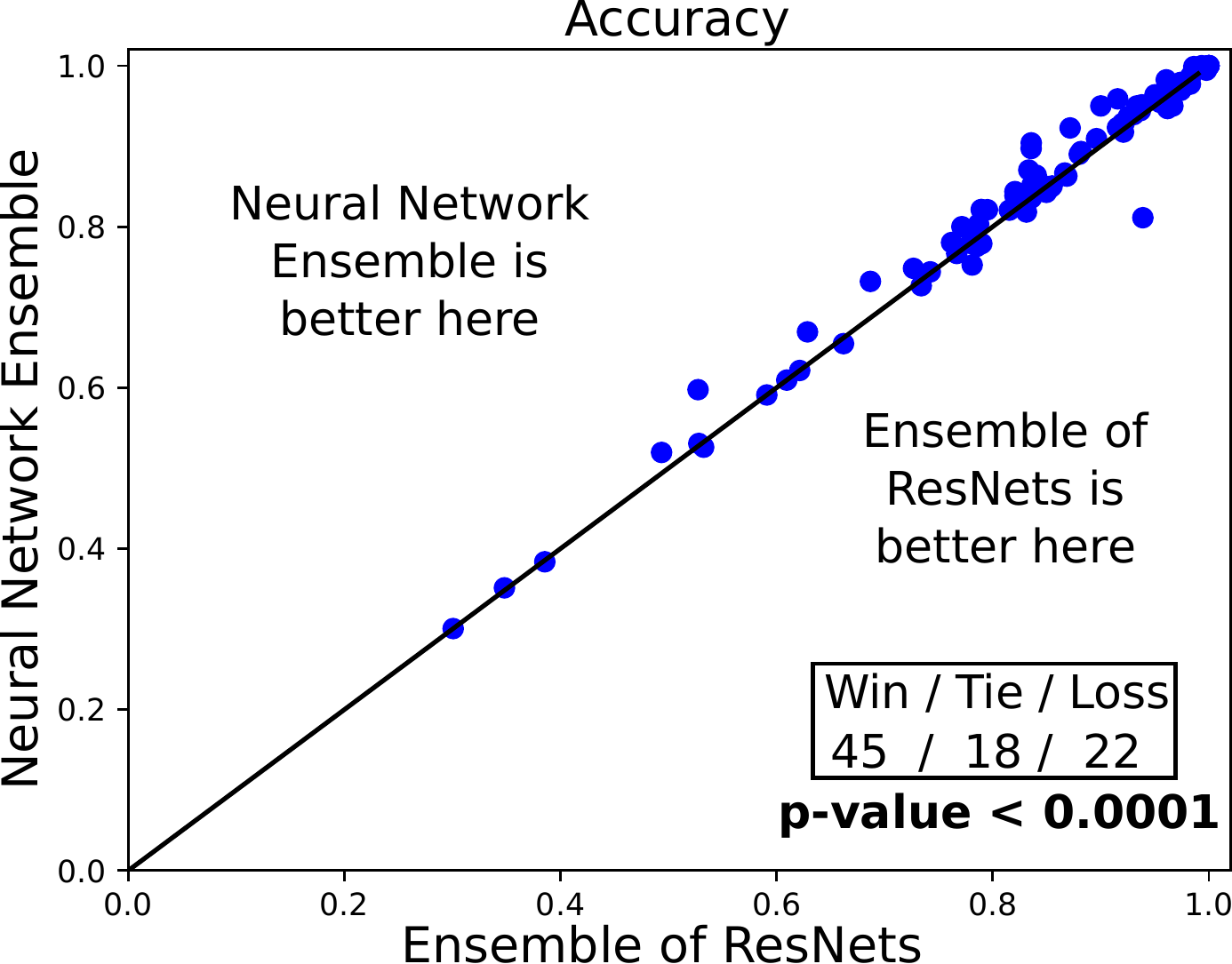}
    \caption{The Neural Network Ensemble (NNE) composed of ResNet, FCN and Encoder is significantly better than an ensemble of pure ResNets.}
    \label{fig:resnet-vs-nne}
\end{figure}

The results in the previous section, suggest that choosing carefully the classifiers in the pool would yield to a better ensemble. 
Therefore, we construct a Neural Network Ensemble (NNE) comprised solely of ResNet, FCN and Encoder. 
These three architectures were the only ones to yield significantly different results when a homogenized ensemble was adopted (Figure~\ref{fig:cd-diagram-all}).
Further investigations suggested that FCN performs better than ResNet on electrocardiography datasets~\cite{IsmailFawaz2018deep}, which would motivate researchers to combine these two classifiers in order to have a robust algorithm that improves the accuracy over the whole datasets. 
However, for small datasets such as DiatomSizeReduction, both FCN and ResNet overfitted the dataset very easily with 30\% test accuracy~\cite{IsmailFawaz2018data}, whereas Encoder managed to achieve very good performance with a 92\% accuracy, therefore implying a combination of ResNet, FCN and Encoder would yield to better accuracy on a various range of TSC datasets. 
Figure~\ref{fig:resnet-vs-nne} shows how NNE is able to outperform an ensemble of pure ResNets with 45 wins and 18 ties on 85 datasets from the archive. 
We believe that the combination of an FCN with ResNet and Encoder, enables the classifier to benefit respectively from the residual linear connections and the attention mechanism. 

\begin{figure}
    \centering
    \includegraphics[width=\linewidth]{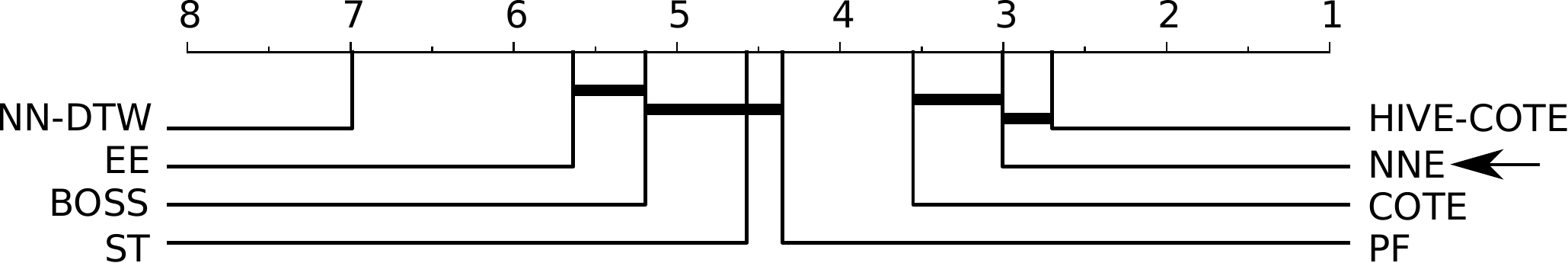}
    \caption{Critical difference diagram showing the pairwise statistical comparison of current state-of-the-art algorithms with the Neural Network Ensemble (NNE) added to the pool.}
    \label{fig:cd-diagram-bake-off}
\end{figure}

To further understand how NNE is performing with respect to current state-of-the-art TSC algorithms, we illustrate in Figure~\ref{fig:cd-diagram-bake-off} a critical difference diagram containing NNE and seven other non deep learning based classiifers: 
(1) NN-DTW corresponds to the nearest neighbor coupled with the Dynamic Time Warping distance; 
(2) EE is an ensemble of nearest neighbor classifiers with eleven elastic distances; 
(3) BOSS corresponds to the ensemble Bag-of-SFA-Symbols; 
(4) ST is another ensemble of off-the-shelf classifiers computed over the Shapelet Transform data domain; 
(5) PF or Proximity Forest is an ensemble of decision trees coupled with eleven elastic distances; 
finally (6) COTE and (7) HIVE-COTE are two ensembles of respectively 35 and 37 classifiers using multiple data transformation techniques. 
The results for these classifiers were taken from~\cite{bagnall2017the} except for PF whose results were taken from the original paper~\cite{lucas2018proximity}.  
Figure~\ref{fig:cd-diagram-bake-off} clearly shows how our NNE is able to reach state-of-the-art performance for TSC, suggesting that CNNs are able to extract one dimensional discriminant features useful for classification in an end-to-end manner, as opposed to other hand-engineered features used by HIVE-COTE such as the Discrete Fourier Transform, DTW features and the Shapelet Transform.        

\subsection{Ensembling fine-tuned models}
\begin{figure}
    \centering
    \includegraphics[width=0.8\linewidth]{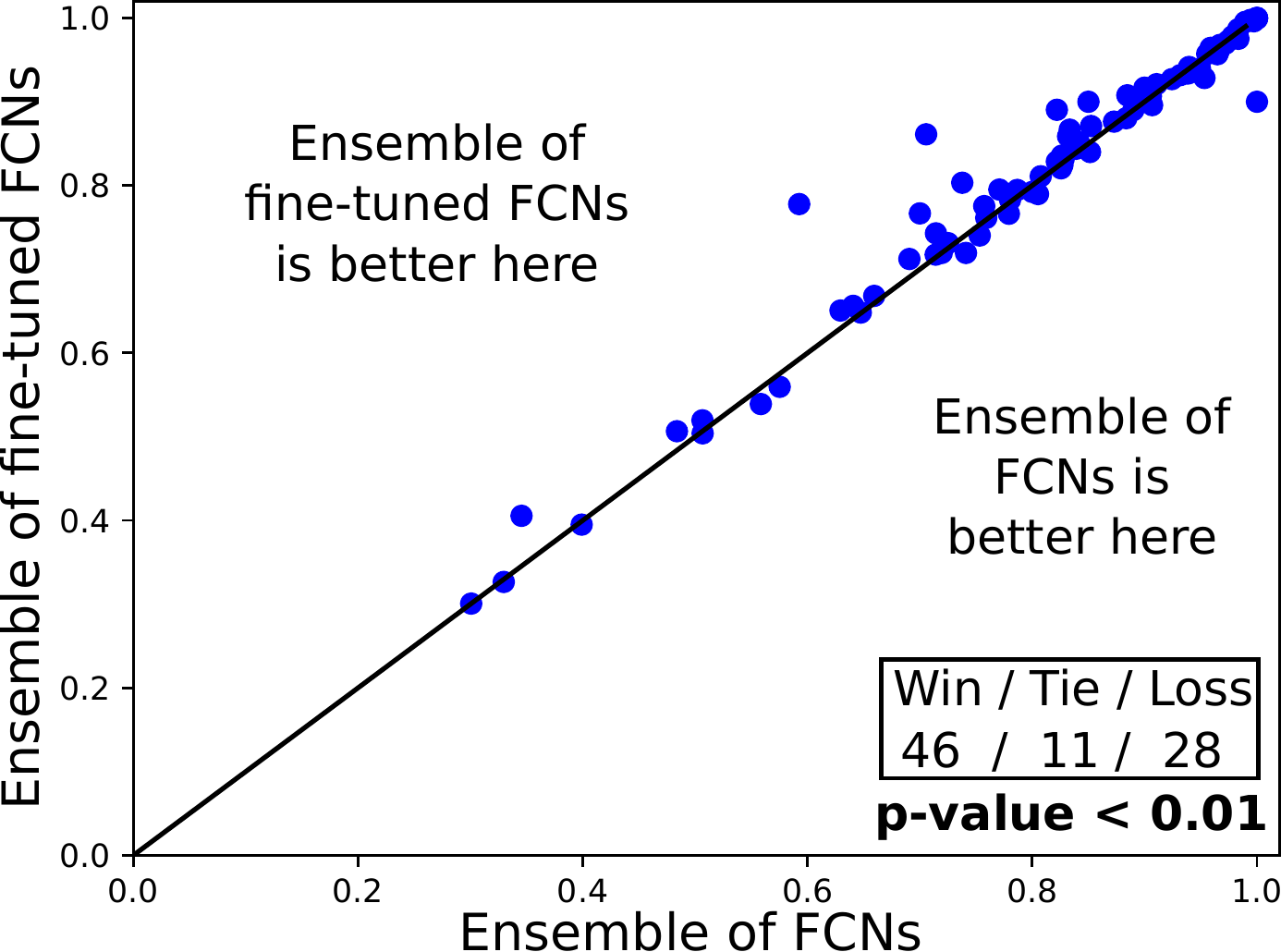}
    \caption{Ensembling fine-tuned models is significantly better than ensembling randomly initialized FCN models that are trained from scratch.}
    \label{fig:fcn-vs-transfer}
\end{figure}

Figure~\ref{fig:fcn-vs-transfer} shows that ensembling fine-tuned FCNs is significantly better than ensembling randomly initialized FCN models that are trained from scratch. 
However, this transfer learning based ensemble did not manage to outperform ResNets' ensemble nor NNE. 
These results show that the choice of architecture is very crucial and suggest that an ensemble of transferred ResNets would demonstrate even better performance than an ensemble of pure ResNets or NNE.   

\section{Conclusion}
In this paper, we showed how ensembling deep neural networks can achieve state-of-the-art performance for time series classification. 
We showed that it would be almost always beneficial to ensemble randomly initialized models rather than choosing one trained neural network out of the ensemble. 
Finally, we investigated an ensemble of transferred deep CNNs to demonstrate even better performance than ensembling randomly initialized networks.
In the future, we would like to consider a meta-learning approach where the output logistics of individual deep learning models are fed to a meta-network that learns to map these inputs to the correct prediction. 

\section*{ACKNOWLEDGMENT}
The authors would like to thank the providers of the UCR/UEA benchmark datasets, as well as NVIDIA Corporation for the GPU Grant and the M\'esocentre of Strasbourg for providing access to the GPU cluster. 

\bibliographystyle{IEEEtran}
\bibliography{biblio}

\end{document}